\documentclass{svproc}
\usepackage{url}

\usepackage{graphicx}
\usepackage{float}

\begin{document}
\mainmatter              
\title{A qualitative investigation of optical flow algorithms for video denoising }
\titlerunning{Optical flow for denoising}  
%
\author{Hannes Fassold}
%
\authorrunning{Anonymous et al.} 
%
\tocauthor{Anonymous author}
\institute{JOANNEUM RESEARCH - DIGITAL, Graz, Austria \\
\email{hannes.fassold@joanneum.at}
}

\maketitle              

\begin{abstract}
A good optical flow estimation is crucial in many video analysis and restoration algorithms employed in application fields like media industry, industrial inspection and automotive. In this work, we investigate how well optical flow algorithms perform qualitatively when integrated into a state of the art video denoising algorithm. Both classic optical flow algorithms (e.g. TV-L1) as well as recent deep learning based algorithm (like RAFT or BMBC) will be taken into account. For the qualitative investigation, we will employ realistic content with challenging characteristic (noisy content, large motion etc.) instead of the standard images used in most publications.
\end{abstract}
\section{Introduction}
\label{sec:intro}
Optical flow, the calculation of the pixel-wise motion between two images, is a crucial step in many video analysis and restoration algorithms. These video analysis \& restoration algorithms subsequently are employed in many application fields, ranging from the media industry (e.g. digital remastering / film restoration) to industry applications (defect detection) and autonomous driving (obstacle avoidance, path planning etc.) and many more. 

Due to its practical importance, a lot of work on fast and robust optical flow algorithms has been published, both classic methods (e.g using variational calculus) as well as modern deep learning based methods. Usually, in these publications the optical flow methods are compared with metrics like average endpoint or average interpolation error on a standard test dataset. Although these metric can serve as a rough guide for choosing a proper optical flow algorithm, they are not able to predict how well an optical flow algorithm performs when integrated in a certain video analysis algorithm (e.g., in a denoising algorithm) and for content of a certain kind (e.g. content with large motion).

In this work, we want to shed some light on how optical flow algorithms perform in the context of video restoration, specifically for the important task of video denoising. This is a sensible way to go, because optical flow calculation almost never is done as an end in itself, but usually as an intermediate step in a higher-level algorithm. Furthermore, we will employ  instead of the standard test images employed in the literature content which is more challenging with large motion, heavy noise or illumination variations. This kind of content can -- and will -- appear in many application fields. We will not calculate quantitative metrics to compare the algorithms, as quantitative metrics (like PSNR for comparing denoising quality) do not correlate well with the user perception. Instead, we will compare the denoising results employing different optical flow algorithms qualitatively (in a subjective way), mainly based on the amount of corruptions like blurring, motion-compensation artefacts or ghosting appearing in the denoised image.

Regarding related work, to our knowledge there are no publications in the field which cover exactly the same topic as our work, investigating the effect of the optical flow method on the result of a video restoration algorithm. But of course there are several surveys for optical flow calculation. A quite extensive survey of the \emph{classic} (not relying on deep learning) optical flow methods can be found in the work \cite{Fortun2015OpticalFM}. In \cite{Sun2010SecretsOO} a couple of tricks (median filtering, bicubic interpolation etc.) are presented in order to improve the performance of classic optical flow algorithms. Finally, the works of \cite{Savian2020OpticalFE} and \cite{Hur2020OpticalFE} provide a decent overview about recent deep learning based approaches for optical flow calculation.

The work is structured as follows. Section \ref{sec:flow} gives a short introduction of the optical flow algorithms which are compared in the context of video denoising. In section \ref{sec:denoiser} we will describe briefly the state of the art denoising algorithm (which internally uses the various optical flow algorithms for motion compensation) we employ and present the denoising results for a certain kind of content and all investigated optical flow algorithms. Finally, section \ref{sec:conclusion} concludes the work.

\section{Optical flow algorithms}
\label{sec:flow}


In the following, we briefly describe the optical flow algorithms which we investigate in section \ref{sec:denoiser} in the context of motion-compensated video denoising. We take into account both classic methods as well as deep learning based methods. 

The \textbf{TV-L1} optical flow algorithm \cite{Wedel2008AnIA} is a very popular method for optical flow estimation. It employs varitional calculus and minimizes a functional containing a data term using the $L_1$ norm together with a regularization term based on the total variation of the flow. Via the total variation regularization, discontinuities in the flow field can be preserved. Furthermore, a structure-texture decomposition of the input images is done in order to be more robust to brightness variations in the content. We employ a GPU implementation of the algorithm.

The Dense Inverse Search (\textbf{DIS}) optical flow algorithm proposed in \cite{Kroeger2016FastOF} is designed specifically for runtime efficiency. It does an efficient search for patch correspondences ($8x8$ pixel patches are employed) inspired by the inverse compositional image alignment method from \cite{Baker2001EquivalenceAE}. We use the CPU implementation provided in the OpenCV library \footnote{\url{https://github.com/opencv/opencv}}. It adds some useful steps to the original algorithm, like spatial propagation of flow vectors and mean normalization of the patches for better robustness against brighness variations.

The \textbf{NV} optical flow algorithm \footnote{\url{https://developer.nvidia.com/opticalflow-sdk}} is a hardware-accelerated optical flow provided on recent GPUs (Turing or Ampere generation) by NVIDIA. The algorithm is very fast as it is implemented completely in hardware. No information has been given by NVIDIA whether the method is a classic or deep-learning based algorithm, and on which publication the implementation is based on.

The \textbf{RAFT} method proposed in \cite{Teed2020RAFT} builds a deep neural network architecture for optical flow and is currently one of the best optical flow methods, when employing the endpoint error metric, on the Middlebury benchmark dataset \footnote{\url{https://vision.middlebury.edu/flow/eval/results/results-e1.php}}. RAFT extracts per-pixel features, builds a multi-scale 4D correlation volumes for all pairs of pixels, and iteratively updates a flow field through a recurrent unit that performs lookups on the correlation volumes. We employ a GPU implementation of this algorithm in PyTorch.

The \textbf{BMBC} algorithm \cite{Park2020BMBCBM} is actually a \emph{frame interpolation} algorithm, but it can be employed also as an optical flow algorithm because the interpolation timepoint $t$ can be chosen freely (so it can be also $0$ or $1$ which corresponds to one of the two input images). It relies on neural networks for bilateral motion estimation followed by dynamic blend filter networks for generating the interpolated (motion-compensated) image. It is currently one of the best optical flow methods, when employing the interpolation error metric, on the Middlebury benchmark dataset \footnote{\url{https://vision.middlebury.edu/flow/eval/results/results-i1.php}}. We employ a GPU implementation of this algorithm in PyTorch. Note that the method has a complex workflow and is therefore significantly slower (one optical flow calculation takes several seconds) than all other optical flow algorithms presented in this work.

Finally, the \textbf{LIFE} optical flow method \cite{Huang2021LIFELI} is a neural network which is explicitly designed to be robust against content with large motion or large brightness variations. It is based on a RAFT-like neural network and takes advantage of weak supervisions from whole-image camera poses. The training is further regularized with cycle consistency and synthetic dense flows. We employ a GPU implementation of this algorithm in PyTorch.


\section{Video denoising algorithm and experiments}
\label{sec:denoiser}

In the following, we first briefly describe the video denoising algorithm for which we will investigate which impact the usage of a certain optical flow method has on the result of the denoising algorithm. We will then investigate qualitatively the impact for challenging content of different kind (like large motion, heavy noise, significant brightness variations) and try to infer from that the specific strong points and weaknesses of a certain optical flow algorithm.

\subsection{Video denoising algorithm}

We integrate the optical flow algorithms into the HWT video denoising algorithm proposed in \cite{Fassold2015MVA}. It is a state of the art wavelet-based video denoising algorithm and is actually integrated in an significantly extended form (with several additional safeguard mechanisms for better robustness against motion compensation errors) into the DIAMANT restoration software \footnote{\url{https://www.hs-art.com/index.php/solutions/diamant-film}} for film and video restoration. 
The HWT video denoising algorithm consists of two phases. In the first phase (\emph{hybrid wavelet denoising}) a wavelet-based denoising is applied using the novel concept of semi-local shrinkage functions. In the second phase (\emph{temporal fusion}), the result images of the first phase are fused within a certain temporal sliding window in a robust way. In both phases, the neighbor images within the temporal sliding window (we use a sliding window of 3 images) are motion-compensated, by first calculating the motion field between a respective neighbor image and the center image (with one the optical flow methods presented in the last section), followed by warping of the neighbor image with the motion field. A quantitative evaluation of the proposed denoiser algorithm on a realistic test dataset with noise of different coarseness and magnitude shows that the proposed method delivers better results than the state of the art video denoising method CVBM3D. There is also an additional preprocessing phase, which is not described in the paper \cite{Fassold2015MVA}. In the preprocessing phase, an accurate noise estimation is done in the wavelet domain, so a sort of \emph{noise profile} is calculated. The estimated noise profile is utilized then in the first phase as a guide.

It is important to note that the employed video denoising algorithm does the motion compensation (warping of the input frames with the motion field calculated by a certain optical flow algorithm) internally, and this step cannot be switched off. So it cannot take advantage of the motion-compensated input image, for optical flow methods which provide not only the motion field but \emph{also} the motion-compensated (or interpolated) image. This means, that the video denoising algorithm currently does not exploit the full potential of frame interpolation methods like BMBC, which provide a motion-compensated image which is (due to the sophisticated interpolation method) usually significantly better compared with an image warped with a motion field.

\subsection{Experiments}

We performed experiments with the video denoising algorithm for different kinds of video content (with a focus on challenging content) and employing different optical flow algorithms (TV-L1, DIS, NV, RAFT, BMBC, LIFE) for the motion compensation step within the denoising process. For each kind of content, we discuss how well the optical flow algorithms perform and show a prototypical result in one figure.

\begin{figure}[H]
\vspace{0.5cm}
\centering
    \includegraphics[width = 0.95 \textwidth]{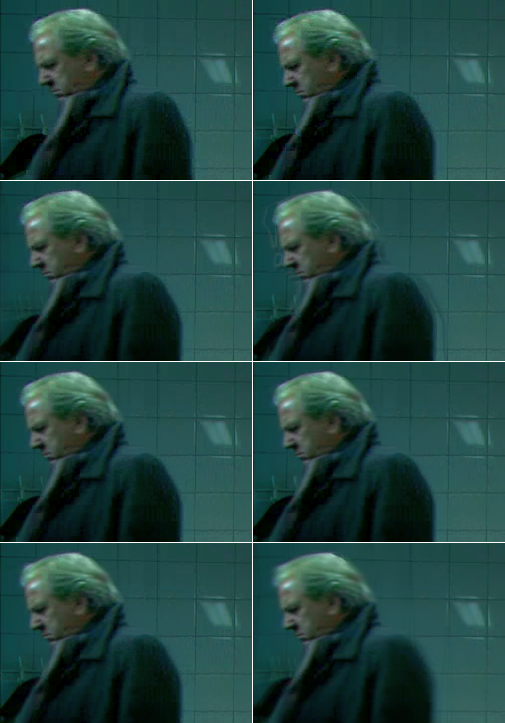}
    \caption{Denoiser result for \emph{normal content with local motion}, employing different optical flow algorithms. Left column (from top to bottom): Original image (previous frame), TV-L1, DIS, NV. Right column (from top to bottom): Original image (current frame), RAFT, BMBC, LIFE.}
    \label{fig:comparison_normal_content_local_motion}
\end{figure}

\textbf{Normal content with local motion.} This is typical content which is often occurring in media. The background of the scene is static, and there is some amount of local motion (e.g. a person moving in the foreground). One can see such content in Figure \ref{fig:comparison_normal_content_local_motion}. On top of the figure are two original images (previous and current frame), and the denoising results (using a certain optical flow algorithm) are shown below. The performance of the TV-L1 method is generally good for this type of content, but some ghosting artifacts are visible on the hooks in the background (to the left of the person). Ghosting artefacts occur when the motion-compensated images do not overlap perfectly (due to non-perfect motion field calculated by the optical flow algorithm). The DIS and NV algorithm seem to be slightly better in that regard, as there are no ghosting artefacts (or other artefacts) observable in the denoising result for them. The RAFT algorithm does not show ghosting artefacts on the hooks, but it shows severe halo artefacts around the moving person. The presence of sever halo artefacts around moving objects indicates that the RAFT algorithm has deficiencies in image areas where occlusions or disocclusions occur. Of course, these areas are difficult to handle for every optical flow algorithm as in these areas no match exists between the previous and current input image. But other optical flow algorithms (e.g. TV-L1 or DIS) set the motion field in these areas so that the motion-compensated (warped) image content in these areas is more similar to the reference image, which has the positive effect of decreasing the severeness of halo artefacts. The BMBC algorithm performs well, no ghosting artefacts or halo artefacts are visible. Finally, the LIFE algorithm shows significant ghosting artefacts in the background, on the hooks to the left as well as on the tiles.

\begin{figure}[H]
\vspace{0.5cm}
\centering
    \includegraphics[width = 0.95 \textwidth]{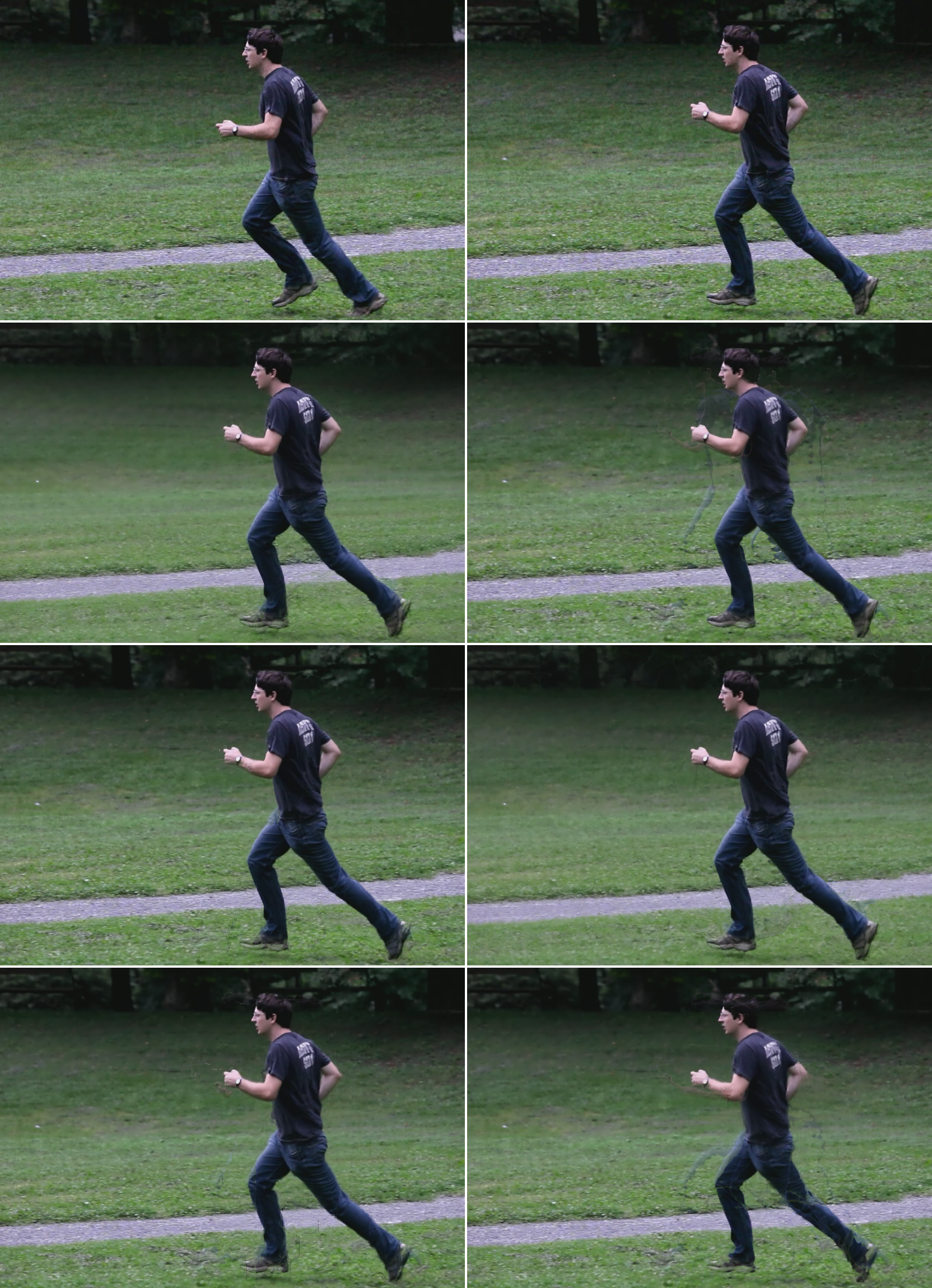}
    \caption{Denoiser result for \emph{content with large motion}, employing different optical flow algorithms. Left column (from top to bottom): Original image (previous frame), TV-L1, DIS, NV. Right column (from top to bottom): Original image (current frame), RAFT, BMBC, LIFE.}
    \label{fig:comparison_content_fast_motion}
\end{figure}

\textbf{Content with large motion.} Content with large motion occurs e.g when broadcasting sport events (fast camera pans etc.). An example is shown in Figure \ref{fig:comparison_content_fast_motion}, where a person is running. The background areas (grass) show a high amount of motion, as the camera is focusing on the running person in the foreground. The TV-L1 method works well for the moving foreground object (the person), no artefacts are visible. But the grass in the background is blurred out significantly. It seems that the TV-L1 algorithm is not able the calculate the optical flow properly for content when the amount of motion is too large. Due to this, in the denoiser preprocessing phase the noise is over-estimated, which leads to an overly aggressive denoising (causing the blurring) in the first phase of the HWT denoising algorithm. The DIS and NV algorithm do not show blurring artifacts in the background regions, but a few ghosting and edge-like artefacts near the left hand and right foot can be seen. The RAFT algorithm seems to be robust against large motion, but shows (again) significant halo artefacts near the outline of the person. The BMBC algorithm copes also well with the large motion, but has some minor ghosting artefacts near the right foot. The LIFE algorithm is able to handle the large motion in the background, but the motion compensation of the foreground person is quite bad, as is evident from the blur, ghosting and edge-like artefacts appearing on the person and around the moving person.

\begin{figure}[H]
\vspace{0.5cm}
\centering
    \includegraphics[width = 0.95 \textwidth]{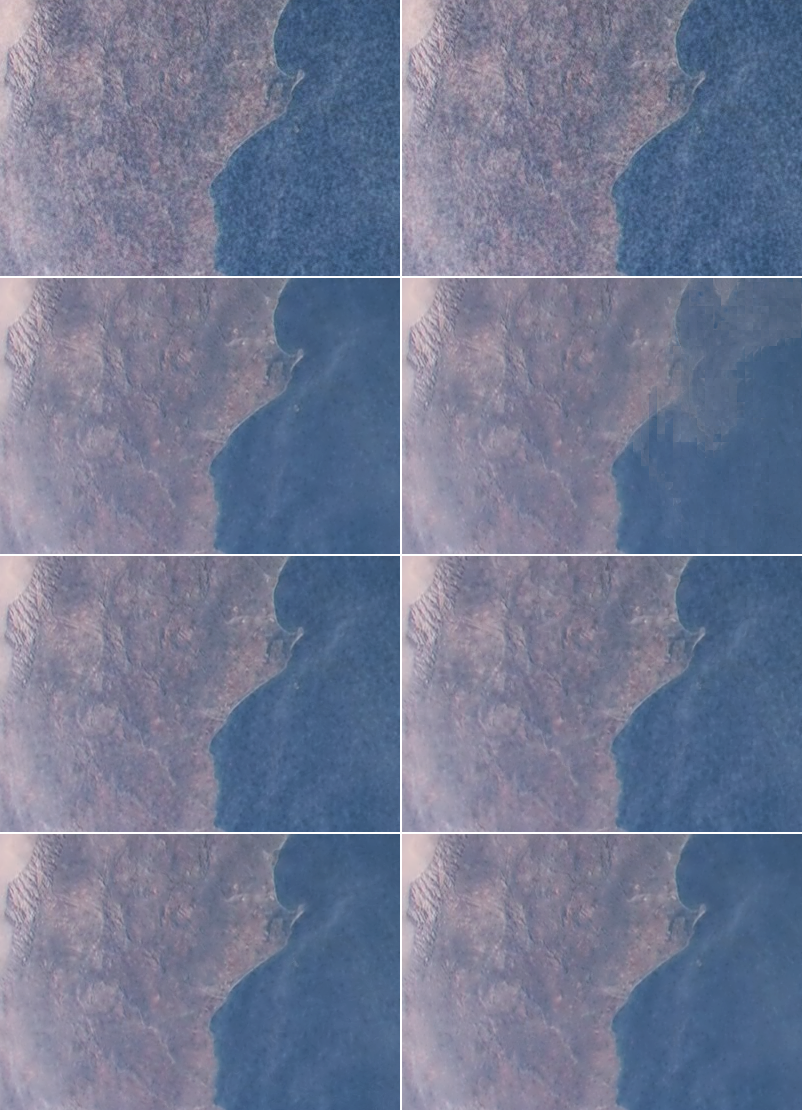}
    \caption{Denoiser result for \emph{content with heavy noise}, employing different optical flow algorithms. Left column (from top to bottom): Original image (previous frame), TV-L1, DIS, NV. Right column (from top to bottom): Original image (current frame), RAFT, BMBC, LIFE.}
    \label{fig:comparison_content_heavy_noise}
\end{figure}

\textbf{Content with heavy noise.} Content can be affected by strong noise, with a very pronounced coarse pattern, when shooting in low-light situations (e.g. night shots). Furthermore, archive film content can be also quite grainy, especially when shot on 16 mm film. In Figure \ref{fig:comparison_content_heavy_noise} synthetically generated noise with a very coarse pattern has been added to clean content with the texture synthesis method given in \cite{Schallauer2006}. The scene itself is nearly static.
For TV-L1, NV and LIFE the result looks fine, nearly all of the noise got eliminated without significant blurring or other artefacts. The results for DIS and BMBC both show still some amount of residual noise, so the denoiser was not successful in getting rid of all of the noise. The most likely reason for this might be that both algorithms are 'matching' the noise pattern, which means that they seem to erroneously  mistake the coarse noise with image structure. Note that in contrast to TV-L1 (which is pixel-based), these methods are operating on patches or CNN features. So they operate on a spatially larger scale than TV-L1, which might explain why they  mistake the coarse noise for actual image structure. For RAFT, the result shows severe motion-compensation artefacts on the upper-right area of the image.

\begin{figure}[H]
\vspace{0.5cm}
\centering
    \includegraphics[width = 0.95 \textwidth]{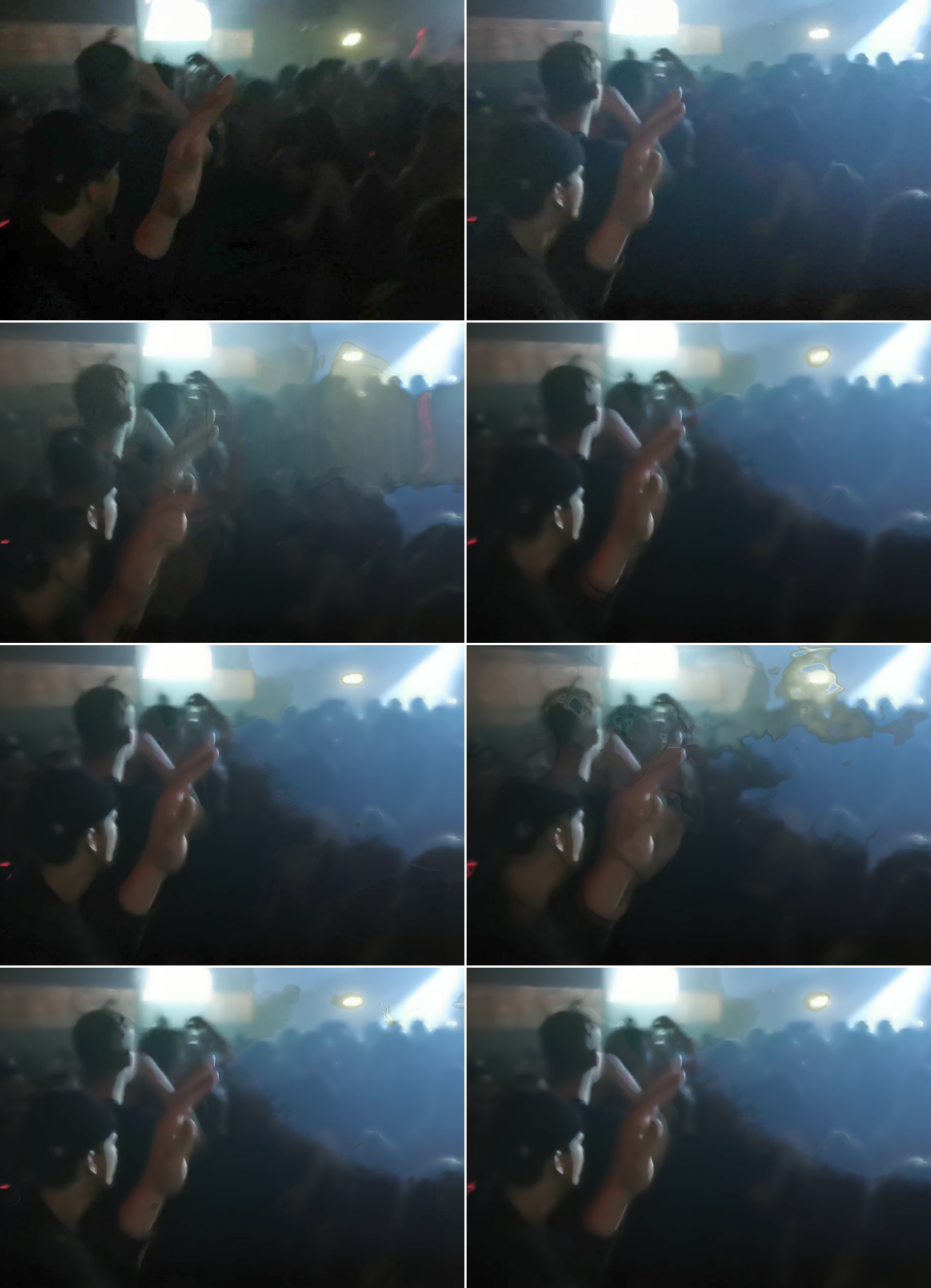}
    \caption{Denoiser result for \emph{content with strong flicker}, employing different optical flow algorithms. Left column (from top to bottom): Original image (previous frame), TV-L1, DIS, NV. Right column (from top to bottom): Original image (current frame), RAFT, BMBC, LIFE.}
    \label{fig:comparison_content_strong_flicker}
\end{figure}

\textbf{Content with strong flicker (brightness variations).} Large brightness variations can occur e.g. in concert settings or when the camera is not synchronized with an artificial light source. In Figure \ref{fig:comparison_content_strong_flicker} an example is given (concert scene). One can see that all results show some cloud-like motion-compensation artefacts, but the amount of artefacts differ. The highest amount of artefacts is visible for the TV-L1 and BMBC method. So it seems that even the structure-texture decomposition done in TV-L1 optical flow (whose aim is to make the method more robust against brightness**variations) does not make the algorithm fully invariant against flicker. The DIS, NV and LIFE algorithm seem to be quite robust against brightness variations, showing only a medium amount of cloud-like artefacts. For the DIS algorithm the reason is clear, as it does internally a mean-normalization of the patches. 

\section{Conclusion}
\label{sec:conclusion}

We did a qualitative investigation how  well  optical  flow  algorithms  perform when integrated into a state of the art video denoising algorithm. Both classic optical flow algorithms as well as recent deep learning based algorithms have been taken into account, and realistic content with challenging  characteristic  (noisy  content,  large  motion  etc.) has been used. From the qualitative tests, no clear "winner" can be determined. But one conclusion is that newer deep learning based optical flow algorithm (especially RAFT) are not necessarily better than the classic methods like TV-L1 or NV. This indicates also that there is a gap between quantitative measures of optical flow performance (like endpoint error - where RAFT is currently one of the top performing algorithms) and a qualitative evaluation in a certain application context. There is clearly a need for more realistic  quantitative measures for the faithful evaluation of optical flow algorithms. 

\section*{Acknowledgments}

We want to thank VRT for providing the witse test video for research and development purposes. This work was supported by European Union´s Horizon 2020 research and innovation programme under grant number 951911 - AI4Media.


%

%
%
\bibliographystyle{bibtex/splncs03_unsrt}
\bibliography{references}

\end{document}